\documentclass[11pt]{article}

\usepackage{acl}
\usepackage{multirow}
\usepackage{times}
\usepackage{latexsym,amsmath}
\usepackage[T1]{fontenc}
\usepackage[utf8]{inputenc}
\usepackage{microtype}
\usepackage{graphicx}
\usepackage{xcolor,soulutf8}
\usepackage{booktabs}

\usepackage{enumitem}
\setlist[enumerate]{itemsep=1pt,topsep=2pt}

\usepackage{subcaption}

\newcommand{\greenhl}[1]{{%
\definecolor{foo}{HTML}{CCEBC5}%
\sethlcolor{foo}\hl{#1}%
}}
\newcommand{\otherhl}[1]{{%
\setlength{\fboxsep}{1pt}%
\definecolor{foo}{HTML}{FED9A6}%
\sethlcolor{foo}\fcolorbox{black}{foo}{#1}%
}}

\title{Text normalization for low-resource languages: the case of Ligurian}

\author{Stefano Lusito\thanks{\hspace*{1em}\texttt{stefano.lusito@uibk.ac.at}.}\hspace*{0.3em}\footnotemark[2] \\
Universität Innsbruck, \\
Institut für Romanistik \\\And
  Edoardo Ferrante\thanks{\hspace*{1em}Equal contribution.} \\
  Conseggio pe-o Patrimònio \\
  Linguistico Ligure \\
  \\\And
  Jean Maillard\thanks{\hspace*{1em}Work conducted in a personal capacity, independently of the author's affiliation.} \\
  Meta AI \\
  }

\begin{document}
\maketitle
\begin{abstract}
Text normalization is a crucial technology for low-resource languages which lack rigid spelling conventions or that have undergone multiple spelling reforms. Low-resource text normalization has so far relied upon hand-crafted rules, which are perceived to be more data efficient than neural methods.

In this paper we examine the case of text normalization for Ligurian, an endangered Romance language. We collect 4,394 Ligurian sentences paired with their normalized versions, as well as the first open source monolingual corpus for Ligurian. We show that, in spite of the small amounts of data available, a compact transformer-based model can be trained to achieve very low error rates by the use of backtranslation and appropriate tokenization.

\begin{center}\large\bf Scintexi\end{center}

A normalizzaçion da grafia a l'é unna tecnologia d'importansa primmäia pe-e lengue con pöche resorse che mancan de regole fisse à sto reguardo ò ch'en passæ pe varie reforme de scrittua. Fin à oua st'operaçion, pe-e lengue de sta categoria, a l'é stæta basâ ciù che tutto in sce de regole mecaniche, tegnue pe ciù efficaçe che l'utilizzaçion di metodi neurali.

Inte ste pagine analizzemmo o caxo da normalizzaçion pe-o ligure zeneise, unna lengua romansa à reisego de scentâ. Da unna parte emmo arrecuggeito 4.394 frase in ligure e ê emmo accobbiæ co-a seu verscion in grafia normalizzâ; da l’atra, emmo creou o primmo corpus monolingue pe-o ligure à liçensa averta. Into studio demostremmo che, pe quante i dati à dispoxiçion seggian scarsci, l'é poscibile allenâ un modello compatto basou in sciô transformer pe razzonze di basci tasci d’errô, pe mezo da backtranslation e de unna tokenizzaçion appropiâ.
\end{abstract}

\section{Introduction}

Many recent advances in the field of NLP rely on large-scale textual corpora. Low-resource languages are typically excluded from such developments due to data scarcity issues. To exacerbate the problem, many low-resource languages suffer from noisy data, due to factors such as the absence of rigid spelling conventions or the lack of user-friendly input methods for special characters \cite{participatory,narabizi}. Text normalization can be a crucial tool to address these issues and enable the creation of corpora from noisy sources \cite{africantextnorm}.

Just as importantly, such technologies can be directly applied to speed up orthographic leveling operations in the field of publishing. Orthographic editing may be desirable when republishing older literary works written according to spelling rules which are no longer in use, or to canonicalize variant forms that may be inadvertently employed even within the same manuscript. This work was born out of such editorial needs for the case of Ligurian, an endangered language spoken within the homonymous region of Liguria in Northern Italy and neighboring territories. The orthographic leveling of historical texts in Ligurian has so far been performed by hand \cite[see e.g.][]{ginna}. The approach presented in this paper represents work conducted by members of the Ligurian linguistic community aiming to largely automate this tedious and time-consuming task with a neural model.

The application of neural methods to text normalization has been limited to high-resource languages \cite{neuralmultitasknorm,shijie1,shijie2,tang-etal-2018-evaluation}. Instead, previous work on low-resource text normalization has relied upon hand-crafted rules \cite{zul,swh,yor}, often perceived as being more data efficient. Indeed the survey by \newcite{histlitrev} recommended the use of substitution lists for low-resource languages and suggested reserving statistical and neural approaches for datasets with at least \textasciitilde10k entries.

In this paper we reassess this belief, looking specifically at the case of Ligurian. We show that, by combining modern data augmentation techniques with a neural model, we are able to achieve average character error rates as low as 2.64 when normalizing texts from a wide range of domains and with a high degree of spelling variation. We achieve this without needing a large-scale annotated corpus: the only resources used are 4.4k unnormalized short sentences (38.1k tokens), manually normalized to produce a parallel corpus (requiring a single native speaker less than five hours of work) and a small unannotated corpus of text (6.7k sentences, 347.5 tokens) which was already in normalized form. We contribute the following:

\begin{enumerate}
  \item A recipe for training a transformer-based text normalization model for an endangered language, showing how backtranslation \cite{bt}, a commonly applied technique in machine translation, can help boost performance in low-resource scenarios. The model and code are released publicly.\footnote{\label{fn1}\url{https://github.com/fleanend/fairseq-text-normalizer}}
  \item A study of the importance of tokenization for neural-based text normalization, showing the effect of varying the vocabulary size for byte-pair encoding \cite{bpe}.
  \item The creation and public release of the first dataset for Ligurian text normalization as well as the first monolingual corpus for Ligurian.\footnote{\url{https://github.com/ConseggioLigure/normalized_ligurian_corpus}}
\end{enumerate}

\section{Background}

The variety of Ligurian we consider in this paper is Genoese, spoken in the capital and neighboring regions \cite{toso-liguria}. It is not only the most widespread dialect in terms of geographical area and number of speakers, but also the only one to possess a written literary tradition that has developed without interruption from the 13\textsuperscript{th} century to the present day \cite{toso-letteratura}.

The spelling system of Genoese has developed over the centuries hand in hand with the evolution of the language itself \cite[p.\,27-32]{toso-letteratura}. A high degree of variation in spelling can still be observed today, due in part to the absence of regulatory bodies and to the lack of keyboard support for several special characters. However, the traditional spelling conventions as formalized by \newcite{toso-grammatica} and recently revised by a group of writers and researchers \cite{acquarone-grafia} are seeing increasing adoption in the publishing and academic worlds \cite{dizToso,gephras,udc,dizLusito}.

In this work, we train models capable of normalizing a variety of Ligurian texts according to the set of spelling conventions mentioned above.

\section{Data collection}
\label{sec:data}

The texts used for this study were chosen to provide a heterogeneous dataset of variously relevant spellings in Genoese literature. They are:
\begin{enumerate}
  \item 1,000 sentences and short examples extracted from \newcite{Casaccia1876}, the most important Genoese-Italian dictionary of the 19\textsuperscript{th} century.
  \item Two poems by \newcite{piaggio}, the most prolific Genoese poet of the first half of the 19\textsuperscript{th} century (1,240 verses).
  \item Two issues of \emph{O Balilla} (1,108 sentences in total), one of the main Genoese bi-weekly papers of the 19\textsuperscript{th} and 20\textsuperscript{th} century. Typographical errors were particularly frequent in this newspaper, making it a perfect source of data for this study.
  \item Five cantos from \newcite{gazzo}'s Genoese translation of Dante's \emph{Comedy} (1,046 verses).
\end{enumerate}

While the first three sources have a similar spelling model\footnote{See \newcite[p.\,104-114]{boano} for an outline of Casaccia's spelling choices.}, Gazzo adopts an extremely complex spelling system, somewhere between the traditional model and a para-phonetic approach \cite[p.\,173-175]{lusito-gazzo}. Unlike the other data sources, Gazzo's spelling system strongly deviates from contemporary Ligurian spelling and does not reflect current usage or even past adoption beyond the author's own work. We nevertheless include it in our analysis to illustrate the model's ability to adapt to strong outliers.

Each one of the datasets above was manually normalized by a native speaker, leading to four parallel corpora which match unnormalized sentences to their normalized versions. For brevity, in experiments we will be referring to these four datasets as \emph{C}, \emph{P}, \emph{B} and \emph{G} respectively. We use a 70/20/10 training/test/validation split throughout this work.

Additionally, we also use a small monolingual corpus of 6,723 sentences of Ligurian, made up of excerpts from the following sources: \emph{O Staf\^i}, a contemporary magazine devoted to sociopolitical discussions, a novel \cite{lazarillo}, and a dozen articles from Ligurian Wikipedia. These texts are largely already in normalized form, apart from a few simple aspects which were fixed in an automated manner.

The divergences between the different spelling systems of the texts considered in this study are illustrated in figure\;\ref{fig:orthocomparison}, showing the target normalized form at the top.

\begin{figure}[ht]
\centering
\setlength{\tabcolsep}{1pt}
\resizebox{\columnwidth}{!}{\begin{tabular}{cccccc}
\greenhl{Unna} & \greenhl{rondaniña} & \greenhl{affammâ} & \greenhl{a} \greenhl{s’} \greenhl{é} \greenhl{pösâ} \greenhl{in} & \greenhl{sciô} & \greenhl{teito} \greenhl{de} \greenhl{coppi} \\
\midrule
\otherhl{Ûnn-a} & \otherhl{rōndaninn-a} & \greenhl{affammâ} & \greenhl{a} \greenhl{s’} \greenhl{é} \greenhl{pösâ} \greenhl{in} & \otherhl{sciö} & \greenhl{teito} \greenhl{de} \otherhl{cōppi} \\
\otherhl{Ûnn-a} & \otherhl{rondaninn-a} & \greenhl{affammâ} & \greenhl{a} \greenhl{s’} \otherhl{è} \greenhl{pösâ} \greenhl{in} & \otherhl{sciö} & \greenhl{teito} \greenhl{de} \greenhl{coppi} \\
\otherhl{Ûnn-a} & \otherhl{rondaninn-a} & \otherhl{affamâ} & \greenhl{a} \greenhl{s’} \otherhl{è} \greenhl{pösâ} \greenhl{in} & \otherhl{sce-o} & \greenhl{teito} \greenhl{de} \greenhl{coppi} \\
\otherhl{Ûña} & \otherhl{rundaniña} & \otherhl{affammä’} & \greenhl{a} \greenhl{s’} \greenhl{é} \otherhl{pösä} \greenhl{in} & \otherhl{sce} \otherhl{o} & \otherhl{téyto} \greenhl{de} \otherhl{cuppi}
\end{tabular}}
\caption{A sample sentence (\emph{A hungry swallow rests on the tiled roof}) in normalized form (top), compared with how it might have appeared in the unnormalized datasets \textbf{C}, \textbf{P}, \textbf{B} and \textbf{G}.} 
\label{fig:orthocomparison}
\end{figure}

\begin{figure}[t]
\includegraphics[width=\columnwidth]{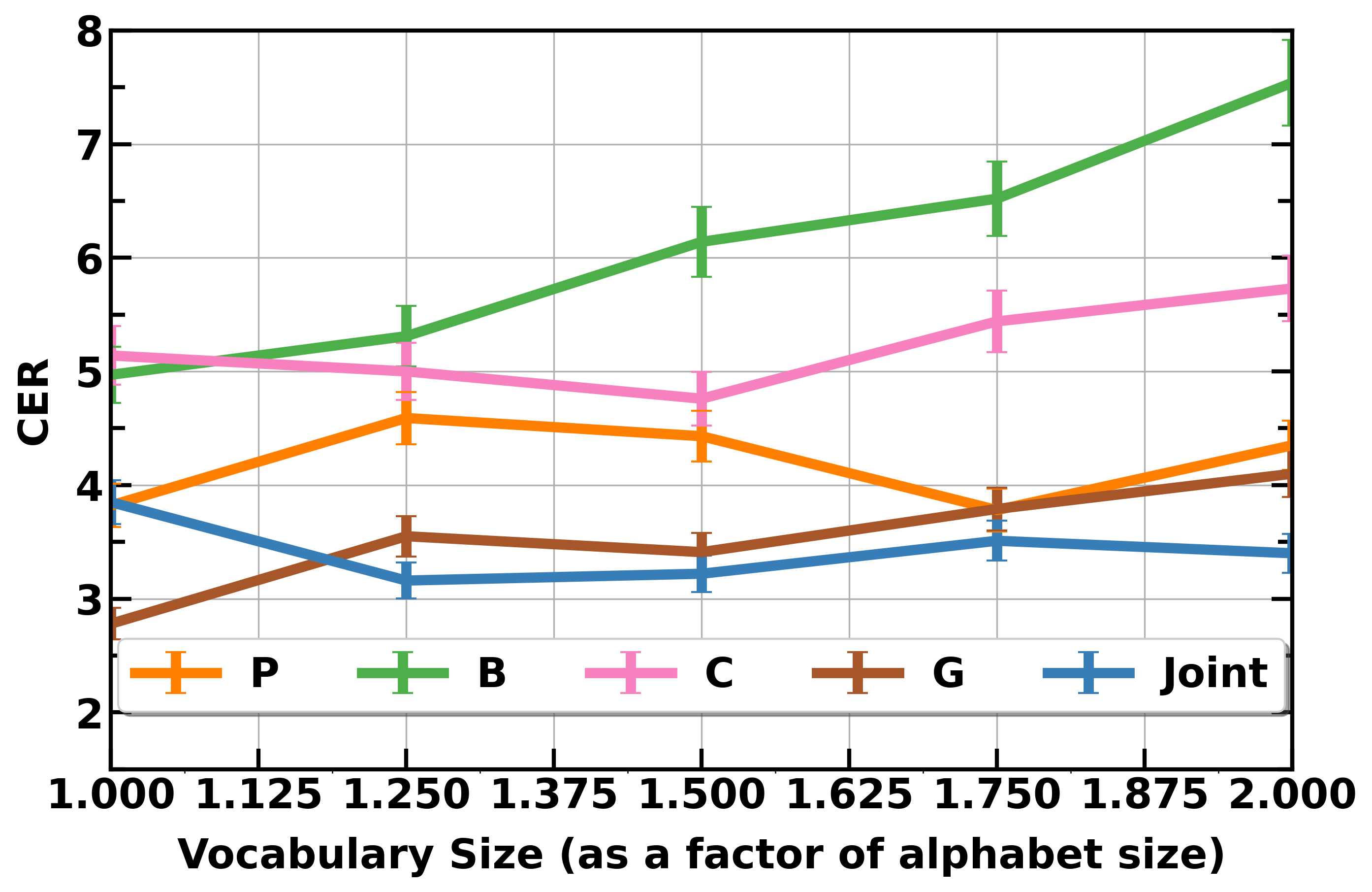}
\caption{Effect of tokenizer vocabulary size on the validation performance of baseline models. We report the average of three runs and standard error.}
\label{fig:tokenizationgraph}
\end{figure}

\section{Experiments}

\subsection{Baseline}
\label{ssec:baseline}

Our baseline is the transformer architecture used to perform historical text normalization by \newcite{shijie1}. After conducting some preliminary experiments to determine if their hyperparameters would be applicable in our highly data scarce setting, we settled upon a much smaller batch size. The resulting architecture, which was trained on \texttt{fairseq} \cite{fairseq}, is made up of 4 encoder and decoder layers with 4 attention heads, embeddings of size 256, a hidden size of 1024, dropout of 0.3 and label smoothing with a factor of 0.1. The model was trained with Adam \cite{adam} with an inverse square root scheduler, a learning rate of $10^{-3}$, 4000 warmup updates, and a batch size of 20.

We train one model per each of the four parallel datasets described in section\;\ref{sec:data}. We then also train one more model on the union of all datasets, with the aim of producing a normalization model capable of working with a wide range of texts.

\subsection{Tokenization}

As is common with transformer-based NLP models we apply subword tokenization to our data \cite{sentencepiece}. We hypothesize that vocabulary size might play an important role in a normalization model's performance. We tested several vocabulary sizes, to check whether the tokenization of digraph and initial, medial or caudal position of a grapheme can help or hinder learning. We measure vocabulary size as a factor of the total alphabet size of a dataset, and experiment with values of 1 (corresponding to a character-level model), 1.25, 1.5, 1.75 and 2. Alphabet sizes for the four datasets are 66, 63 , 50 and 78 for the \emph{P}, \emph{B}, \emph{C} and \emph{G} datasets respectively; their union has size 82.

As shown in figure\;\ref{fig:tokenizationgraph}, we see that, especially for the spelling systems with the least resources, the level of tokenization (determined by the vocabulary size) has a noticeable impact on the model performance. We suspect the optimal number is highly dependent on the number of unigraph $\leftrightarrow$ multigraph alignment pairs in each parallel dataset: having a multigraph tokenized as a single unit should aid in the task of mapping it to a monograph, while having it split up into multiple tokens should make the model's task harder.

We select vocabulary sizes for each setup based on this validation performance and keep them fixed for the remaining experiments.

\subsection{Joint model}

A normalization model is most useful if it can operate on texts regardless of their domain or the nature of their spelling variation, and without the need to tag them as being from specific sources. We attempt to create a universal normalizer by merging all datasets described in section\;\ref{sec:data} and training one more model on this unified parallel corpus.

Table\;\ref{tbl:baseline} compares the results of the joint and dataset-specific models. For reference, we additionally report the performance of a naive ``copy'' baseline, which simply leaves text as-is. We also report the performance on the joint dataset of the dataset-specific models, which consists in sending each sample to the appropriate dataset-specific model depending on its origin.

We find from these results that the joint model still manages to achieve sub-5 CER on all datasets, even outperforming the dataset-specific models in a majority of cases. We believe that concatenating all data helps the model learn the target spelling system better, while consolidating any grapheme normalization rules in common among the sources. On the other hand, the joint model also learns on contrasting evidence for several graphemes, favoring thus the spelling systems with the most features in common (\emph{B} and \emph{C}) and punishing the most eclectic ones. This is particularly evident for the case of \emph{G} which, as already noted in section\;\ref{sec:data}, is by far the most complex to deal with.

\begin{table}[t]
  \centering
  \begin{tabular}{lccccc}
    \toprule
    & \multicolumn{5}{c}{Dataset} \\
    & \emph{P} & \emph{B} & \emph{C} & \emph{G} & Joint \\
    \cmidrule{2-6}
    Specific & 3.78 &  4.97 &  5.14 & \textbf{2.78} & 4.42 \\
    Joint & \textbf{2.57} & \textbf{3.29} &  \textbf{2.28} & 4.32 & \textbf{3.16} \\
    \midrule
    \emph{Copy} & 9.54 & 8.24 & 21.05 & 14.59 & 12.76 \\
    \bottomrule
  \end{tabular}
  \caption{Test set performance (CER) of the dataset-specific and joint models of section\;\ref{ssec:baseline}, compared to a naive ``copy'' baseline. Best of three runs based on validation performance.}\label{tbl:baseline}.
\end{table}

\subsection{``Backnormalization''}
\label{ssec:bn}

Backtranslation \cite{bt} is an extremely effective approach used in machine translation to benefit from unannotated, monolingual data \cite{bt-scale}. We adapt this method to the task of text normalization as follows. We train an additional set of dataset-specific baseline models as in section\;\ref{ssec:baseline}, but this time on the reverse task, going from normalized to unnormalized text (``backnormalization'' for brevity). We then take our unannotated corpus, which is already in normalized form, and run it through these four backnormalization models. The result of this procedure are four new pseudo-parallel corpora, the target side being our unannotated corpus, and the source side being our backnormalization models' attempts at reconstructing how this text might have been written in unnormalized form, with the spelling variations typical of the four sources of data under study in this paper.

These backnormalized datasets were then used in addition to our original training data. Due to the noisy nature of backnormalized data, which is half-synthetic, in training we upsample the original parallel datasets with a factor of $\left \lfloor{N_\text{backnormalized}/N_\text{original}}\right \rfloor$ (where $N$ is the length in tokens of a corpus). This is so that in training the model would learn from an equal number of human-normalized and backnormalized data.

To study the effectiveness of backnormalization as well as the effect of the size of the unannotated corpus, we repeat this experiment twice. First using a 3.7k-sized portion of the unannotated corpus, then using the full dataset. The results in table\;\ref{tbl:bn} confirm that backnormalization is indeed effective, showing a noticeable reduction in character error rate overall for the models trained on the augmented data. We also a fairly clear trend of improvement in the error rate when more unannotated data is used via backnormalization.

One notable case is performance on dataset \emph{G}, which shows mild improvement in the dataset-specific models but degradation for the joint model. As previously discussed, the spelling system of this dataset represents a clear outlier, deviating strongly from actual contemporary usage. We note that in any kind of real-world use case such type of data would be considered out-of-scope.

\begin{table}[t]
  \centering
  \begin{tabular}{lccccc}
    \toprule
    & \multicolumn{5}{c}{Dataset} \\
    & \emph{P} & \emph{B} & \emph{C} & \emph{G} & Joint \\
    \cmidrule{2-6}
    Specific+BN & 2.26 &  4.95 &  2.32 & \textbf{2.39} & 3.11 \\
    Joint+BN & 2.80 & 2.96 &  1.67 & 4.42 & 2.98 \\
    \midrule
    Specific+BN' & 2.28 &  2.45 &  2.52 & 2.44 & \textbf{2.47} \\
    Joint+BN' & \textbf{1.91} & \textbf{2.38} &  \textbf{1.36} & 4.55 & 2.64 \\
    \bottomrule
  \end{tabular}
  \caption{Test set performance (CER) of the dataset-specific and joint models when augmenting training data via ``backnormalization'' (section\;\ref{ssec:bn}) of 3.7k sentences (+BN) and all 6.7k sentences (+BN') from an unannotated corpus. Best of three runs based on validation performance.}\label{tbl:bn}.
\end{table}

\section{Conclusion}

In this paper we have tackled the issue of text normalization for Ligurian, an endangered low-resource language. We did so by collecting and releasing a dataset of 4,394 Ligurian sentences in different spelling systems paired with normalized versions. We further gathered and released the first open source digital monolingual corpus of contemporary Ligurian, consisting of 6,723 sentences, and showed its potential despite its modest size.

We have shown that in low-resource settings a compact transformer-based model with the appropriate choice of hyperparameters and tokenization, combined with ``backnormalization'', can achieve CER under 3 points on average, even when the model is given no information on the provenance and spelling conventions of the source text. By varying the size of the corpus used with backnormalization we have further shown that performance is likely to improve even further, should more unannotated data be collected.

There are multiple practical applications of such a model. First of all, it could allow the general public -- which tends to experiences difficulties entering Ligurian characters by means of the Italian keyboard -- to publish texts in a relatively uniform spelling. Second, on the editorial side, it would greatly speed up orthographic leveling operations, not just of contemporary material but also for the republishing of older literary works written according to spelling rules which are no longer in use.
Finally, the normalization of corpora with noisy spelling is especially important for a low-resource language, as it would make more data available for downstream tasks such as language modelling and machine translation which are reliant upon the availability of large corpora.

In conclusion, we consider neural text normalization combined with backnormalization to be a particularly useful tool to promote the preservation as well as the revival of the Ligurian language. The present work only focuses on the case of Ligurian. However, due to the low amount of data needed to train a normalization system, we hope that other researchers and members of language communities in analogous situations will be able to adapt and apply this approach.

\bibliography{main}
\bibliographystyle{acl_natbib}

\appendix

\end{document}